\documentclass[letterpaper, 10 pt, conference]{ieeeconf}  

\usepackage{times}
\usepackage{caption}
\usepackage{microtype}
\usepackage[table]{xcolor}
\usepackage{graphicx}
\usepackage{booktabs} 
\usepackage{tikz}
\usepackage{textcomp}
\usepackage{float}
\usepackage{algorithm}
\usepackage{algorithmic} 
\usepackage{multicol}
\usepackage{multirow}
\usepackage[bookmarks=true]{hyperref}
\usepackage{subfigure}
\usepackage{amsmath} 
\usepackage{enumerate}
\usepackage{amssymb}
\usepackage{amsbsy}
\usepackage{import}

\IEEEoverridecommandlockouts                              

\title{\LARGE \bf
Deep Reinforcement Learning Based on Local GNN for Goal-Conditioned Deformable Object Rearranging
}

\author{Yuhong Deng$^{1, 2}$, Chongkun Xia$^{2}$, Xueqian Wang$^{2}$ and Lipeng Chen$^{1,\dagger}$ 
\thanks{ ${1}$ Tencent Robotics X Lab, Shenzhen, China. 
{\tt\small \{francisdeng, lipengchen\}@tencent.com}}
\thanks{ ${2}$ The Center for  Intelligent  Control  and  Telescience,  Tsinghua  Shenzhen  International Graduate School, Shenzhen, China 
{\tt\small \{xiachongkun, wang.xq\}@sz.tsinghua.edu.cn}}
\thanks{ ${\dagger}$ Corresponding author}
}

\begin{document}

\maketitle
\thispagestyle{empty}
\pagestyle{empty}


\begin{abstract}
 Object rearranging is one of the most common deformable manipulation tasks, where the robot needs to rearrange a deformable object into a goal configuration. Previous studies focus on designing an expert system for each specific task by model-based or data-driven approaches and the application scenarios are therefore limited.
 Some research has been attempting to design a general framework to obtain more advanced manipulation capabilities for deformable rearranging tasks, with lots of progress achieved in simulation. However, transferring from simulation to reality is difficult due to the limitation of the end-to-end CNN architecture.
 To address these challenges, we design a local GNN (Graph Neural Network) based learning method, which utilizes two representation graphs to encode keypoints detected from images. Self-attention is applied for graph updating and cross-attention is applied for generating manipulation actions. Extensive experiments have been conducted to demonstrate that our framework is effective in multiple 1-D (rope, rope ring) and 2-D (cloth) rearranging tasks in simulation and can be easily transferred to a real robot by fine-tuning a keypoint detector. 
  
\end{abstract}


\section{Introduction}
\label{sec:introduction}
Deformable objects can be seen in many automating tasks such as food handling, assistive dressing and the manufacturing, assembly and sorting of garment~\cite{appli_1, appli_2, appli_3}. Rearranging deformable objects is one of the most investigated and fundamental deformable manipulation tasks, where the robot is expected to infer a manipulation sequence to rearrange a deformable object into a goal configuration. 
Different from rigid object manipulation~\cite{swingbot, throw, deng2019deep}, rearranging deformable objects poses two new challenges. The first challenge lies in the high dimensionality of the configuration space~\cite{def_sci}, which makes the observation and representation of deformable objects more complicated for efficient manipulation planning. The second challenge comes from the complex and non-linear dynamics of deformable materials~\cite{graph_dy}, which makes the movements of deformable objects hard to predict.

\begin{figure}[t]
\centerline{\includegraphics[width=1\columnwidth]{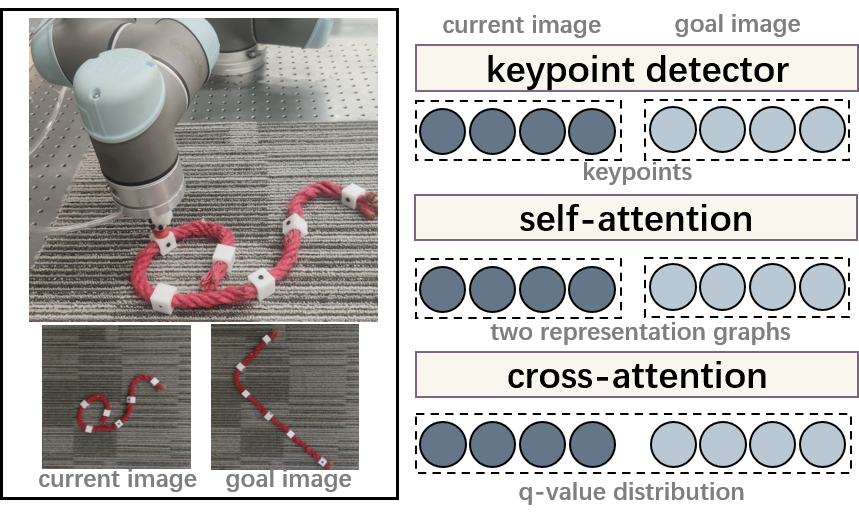}}
\caption{We formulate the rearranging task as a sequence-to-sequence problem: generating the Q-value distribution of actions (represent the probability of picking and placing on certain keypoint pair) from the keypoints detected from current and goal images. We use a local GNN to solve this problem. The Model trained in simulation can be directly transferred to reality with only keypoint detector fine-tune.}
\label{fig:system}
\end{figure}

There are two main solutions for deformable manipulation, including model-based and data-driven approaches. However, most pioneering works tend to provide task-specific models such as folding cloth, untangling rope-knot~\cite{untangling}, and moving ropes in scenes with obstacles~\cite{move}, which limits their generalization in real applications. 
Seita et al.~\cite{transporter_df} proposed a goal-conditioned transporter network, which can be a general learning framework for multiple goal-conditioned deformable rearranging tasks. However, the transporter network is based on an end-to-end CNN architecture for extracting image features, and therefore the training is data-consuming and the algorithm trained in simulation is hard to transfer into reality.

To tackle this problem, we separate the processing of image features from the manipulation planning (Fig.~\ref{fig:system}).
We first propose an efficient configuration representation to improve the generalization of visual features, by encoding keypoints detected from an image into a representation graph.
The graph feature contains more semantic information and therefore can represent the configuration space of a deformable object more efficiently. After obtaining representation graphs of the current and goal configurations for a rearranging task, the two graphs are passed through a self-attention layer (the attention calculation is performed on keypoints from the same image) and the two graphs are updated dynamically. Finally, our framework adopts a cross-attention layer (the attention calculation is performed on keypoints from different images) to infer the manipulation action that can narrow the gap between current and goal configurations from two updated graphs. The self-cross attention strategy can reduce the complexity while ensuring performance, which makes our framework easier to deploy.\par

In addition, existing datasets for deformable rearranging tasks are almost on a specific task, which can not be used to verify the performance of our framework. Therefore, we establish a dataset of rearranging 1-D and 2-D deformable objects, and the results demonstrate the proposed framework is effective and general for goal-conditioned rearranging tasks. Real experiments also demonstrate the enhanced sim-to-real transferability of our framework. The contributions of this paper are  summarized as follows:
\begin{itemize}
    \item [1)]
    We propose a novel configuration representation for deformable manipulation, where keypoints of the deformable object are encoded into a dynamic graph.
    \item [2)]
    We propose an effective and general framework that utilizes self-cross attention for the rearranging of 1-D and 2-D deformable objects.
    \item [3)]
    We narrow the gap between simulation and reality in goal-conditioned deformable object rearranging learning.
\end{itemize}
The rest of this article is organized as follows. The related work is reviewed in Sec. \ref{section:relatedwork}. We introduce our solution framework briefly in Sec. \ref{sec:framework}. In Sec. \ref{sec:learning}, we discuss the detailed learning algorithm in our framework. The dataset we established is presented in Sec. \ref{sec:dataset}. The experimental setup and results are provided in Sec. \ref{sec:experiment}. Finally, we come to the
conclusion of the article in Sec. \ref{sec:conclusion}.

\section{RELATED WORK}
\label{section:relatedwork}
\subsection{Approach for Deformable Object Manipulation}
There are two main approaches:
Model-based approaches rely on physics to define the relationship between the configurations of the deformable object and manipulations. Establishing a data-driven dynamic model that can predict configurations of the deformable object is a new trend for accuracy. Wang et.~\cite{wang_insert} construct a latent space to represent configurations of the deformable object by CIGAN and establish an inverse model. After the data-driving model is established, the sampling-based method~\cite{yan_dynamic}\cite{yan_mpc} is mostly used to get the manipulation related to configurations. However, if the initial configurations and goal configurations object varies considerably, interpolating intermediate configuration will bring an extra effort and cumulative error.\par

Data-driven approaches can be divided into two categories according to the source of the training data: imitation learning and reinforcement learning. In imitation learning, the manipulation task is often formulated as a supervised learning problem where the robot should imitate the observed behaviors. Nair et al.~\cite{nair_human} adopt human demonstrations to solve the multi-step manipulation task. Reinforcement learning agent acquires rearranging skills through robot exploration. Matas et al.~\cite{rl_mata} train an end-to-end RL agent to complete the task of folding a towel and draping a piece of cloth over a hanger. Wu et al.~\cite{rl_condi} make the robot learn a pick-and-place policy for cloth folding and rope straightening by a conditional reinforcement learning method.\par

\subsection{General Framework of Deformable Object Manipulation}
 Designing a learning policy for a specific deformable object manipulation task has been widely investigated. However, the generalization of these algorithms is limited. Designing a general framework for deformable object manipulation has achieved some progress recently. Zeng et al.~\cite{transporter} proposed the transporter network that performs well on several rearranging tasks of rigid objects and Seita et al.\cite{transporter_df} further improved the network and applied it to additional tasks of deformable objects. Lim et al.\cite{lim2021multi} made a more systematic classification of the rearrangement tasks, and achieved better results in sequence-conditioned task learning. Shridhar et al.~\cite{cliport} proposed CLIPort that can produce multi-task policies conditioned on natural language. However, it's hard for these works to transfer to reality for 2 reasons. One reason is using CNN feature directly to represent the deformable object, which brings redundant information for the sparse feature of the deformable object. And the other reason is that end-to-end architecture also highly relies on quantities of data, which is hard to collect in reality. In our model design, the deformable object is represented as a graph, which provides rich semantics including keypoint positions and the topological relation between keypoints. And we use local GNN to process the graph features, which can reduce the size of the model. At the same time, our framework is consist of a keypoint detector (visual feature processing) and a local GNN (manipulation planning). All of these designs make our framework is also effective in reality.
 
\section{Framework}
\label{sec:framework}
\begin{figure}[!t]
\centerline{\includegraphics[width=1\columnwidth]{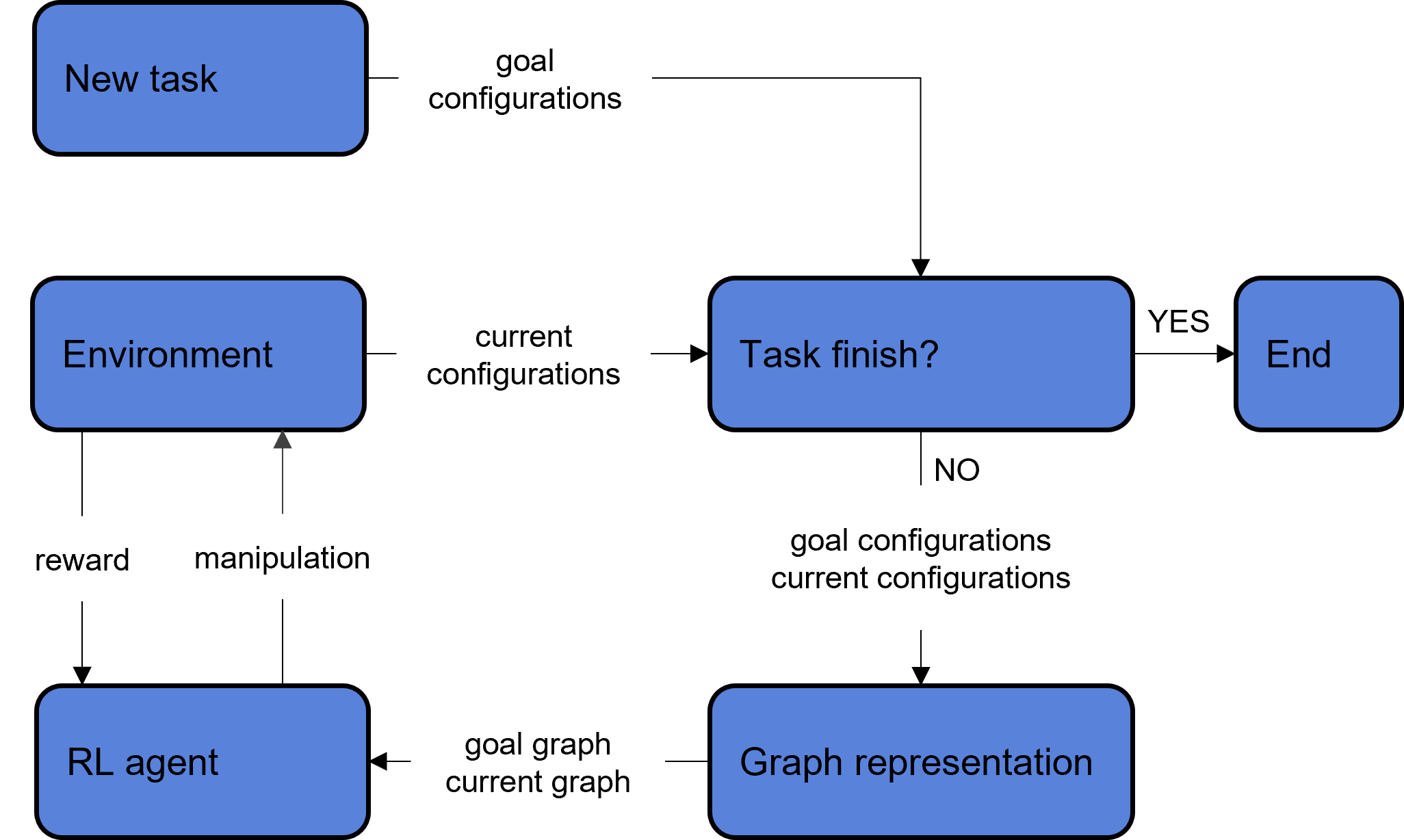}}
\caption{System overview: we proposed an efficient general framework for rearranging the deformable object. The configurations of the deformable are encoded into a graph first. And the RL agent will output a manipulation mapped from the current and goal representation graph to interact with the environment continuously until the task is finished.}
\label{fig:framework}
\end{figure}
To provide a solution framework (Fig.~\ref{fig:framework}) for deformable object rearranging tasks, we need an efficient configuration representation method. The configurations of the deformable object are represented as a graph structure in our framework. We detect keypoints from the image of the deformable object to reduce the dimensionality of the configuration firstly. The keypoint detecting task is formulated as a supervised learning problem. Our keypoint detector outputs the gaussian heatmap transformed from the coordinates of the keypoint instead of directly outputting the coordinates. The training goal of the model is to reconstruct gaussian heatmap instead of keypoint coordinates regression, which has improved the performance of keypoint detecting. These detected keypoints of current and goal images will be used to consist of two representation graphs. Then we adopt an attention mechanism to update the relationship between the vertices in representation graphs.\par
After getting the representation graph, manipulations for rearranging tasks can be generated. Considering that the rearranging task is a complex multi-step manipulation planning task, our goal is to obtain a manipulation that can narrow the gap between current and goal configurations as much as possible instead of obtaining the entire manipulation sequence for rearranging. The agent outputs a manipulation mapped from the current and goal configurations and obtains the new current configurations from the environment observation after the manipulation is taken. The robot will repeat the above processing until the gap between the current and goal configurations is small enough. The manipulation planning task is formulated as a reinforcement learning problem, where the state is the graph representation, the action is the pick and place manipulation and the reward is related to the gap between current and goal configurations. We adopt DQN (Deep Q-Learning) and we use a local Graph Neural Network to approximate the Q-value function.

\section{Learning for Rearranging Tasks}
\label{sec:learning}
\subsection{Learning for Graph Representation}
\label{sec:graph}

\begin{figure}[!t]
\centerline{\includegraphics[width=1\columnwidth]{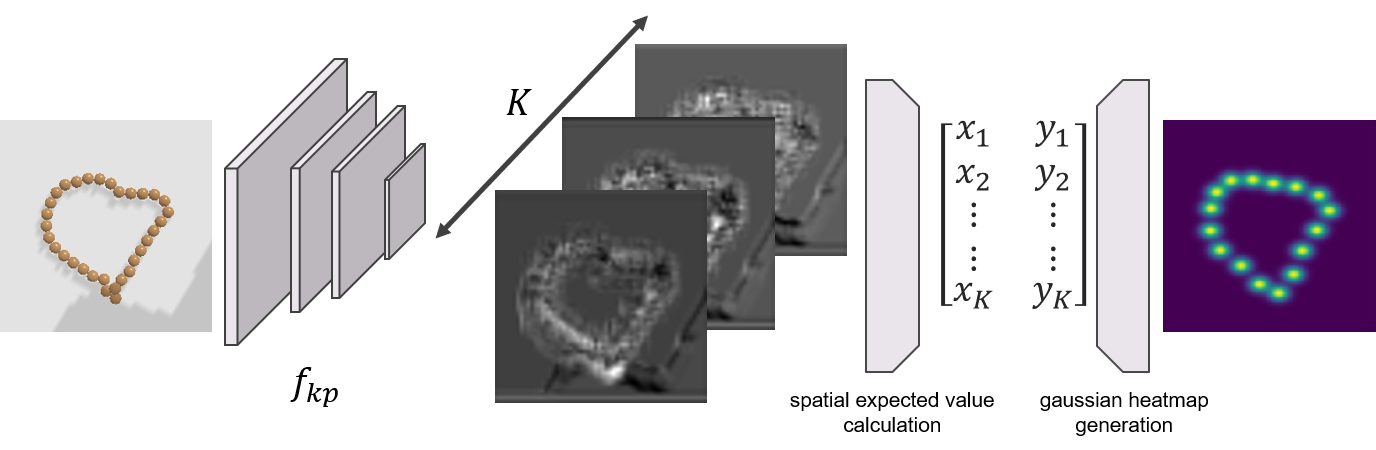}}
\caption{The design of keypoint detector: we design a keypoint detector, the training goal is to reconstruct the gaussian heatmap centered at the coordinates of keypoints.}
\label{fig:detector}
\end{figure}
The motivation of our framework is using a single complete graph whose nodes are the keypoints of the image to represent the deformable object with a large degree of freedom. Keypoint detection is the first step for graph construction. We borrowed the model in~\cite{keypoint} and the design of our keypoint detector model is shown in Fig.~\ref{fig:detector}. Given an image $I \in \mathbb{R}^{H \times W \times 3}$ of a deformable object, the feature CNN $f_{kp}$ extracts a feature map $f_{kp}(I) \in \mathbb{R}^{H^{'} \times W^{'} \times K}$ from $I$ firstly, we use $\Omega$ to denote the image domain ($H\times W$ lattice) and use $\Omega^{'}$ to denote the feature domain ($H^{'}\times W^{'}$ lattice). $K$ heatmaps $\mathcal{H}_c(I;k),c\in \Omega^{'}$ ($k=1,2,3,\dots,K$ ) are obtained in parallel as the channels of $f_{kp}(I)$ ($K$ is the number of keypoints), one heatmap for each keypoint. \par
Second, we can estimate coordinates $(\hat{x_k}^{'},\hat{y_k}^{'})$ of a keypoint by computing the (spatial) expected value of the lattice in each heatmap that has been normalized via spatial softmax:
\begin{equation}
(\hat{x_k}^{'},\hat{y_k}^{'})= \hat{c_k^{'}} = 
\frac{\sum_{c\in \Omega^{'}}c*\exp(\mathcal{H}_c(I;k))}{\sum_{c\in \Omega^{'}}\exp(\mathcal{H}_c(I;k))}.
\label{eq_1}
\end{equation}
The calculation of coordinates is fully-differentiable and can limit the information flow to improve model performance. And we can get the coordinates in image domain $\Omega$ by a direct mapping:
\begin{equation}
\hat{c_k} = (\hat{x_k},\hat{y_k})= (\frac{H\hat{x_k}^{'}}{H^{'}},\frac{W\hat{y_k}^{'}}{W^{'}}).
\label{eq_2}
\end{equation}
Finally, we can get a gaussian heatmap by a gaussian-like function centred at $\hat{c_k}$ with a small fixed
standard deviation $\sigma$:
\begin{equation}
\Phi_c(I;k)=\exp(-\frac{1}{2\sigma^2}\left\|c-\hat{c_k}\right\|^2)
\label{eq_3}
\end{equation}
The end result is a gaussian heatmap $\Phi(I)=\sum_{k=1}^K\Phi(I,k)$, $\Phi(I) \in \mathbb{R}^{H \times W}$, where the location of $K$ maxima is the estimated keypoint coordinates. Our training goal is to make the gaussian heatmap of the estimated keypoint coordinates $C^{'}=\{c_i^{'}=(x_i^{'},y_i^{'})\}_{i=1}^K$ and that of the true keypoint coordinates $C=\{c_i=(x_i,y_i)\}_{i=1}^K$  as close as possible instead of making the keypoint coordinates close directly. It is easy to recover the keypoint coordinates exactly from these heatmaps, this gaussian heatmap representation is equivalent to the 2D coordinates. In addition, regressing the heatmap will make the model easier to train compared to regressing coordinates directly because of a higher feature dimension.\par
\begin{figure}[!t]
\centerline{\includegraphics[width=1\columnwidth]{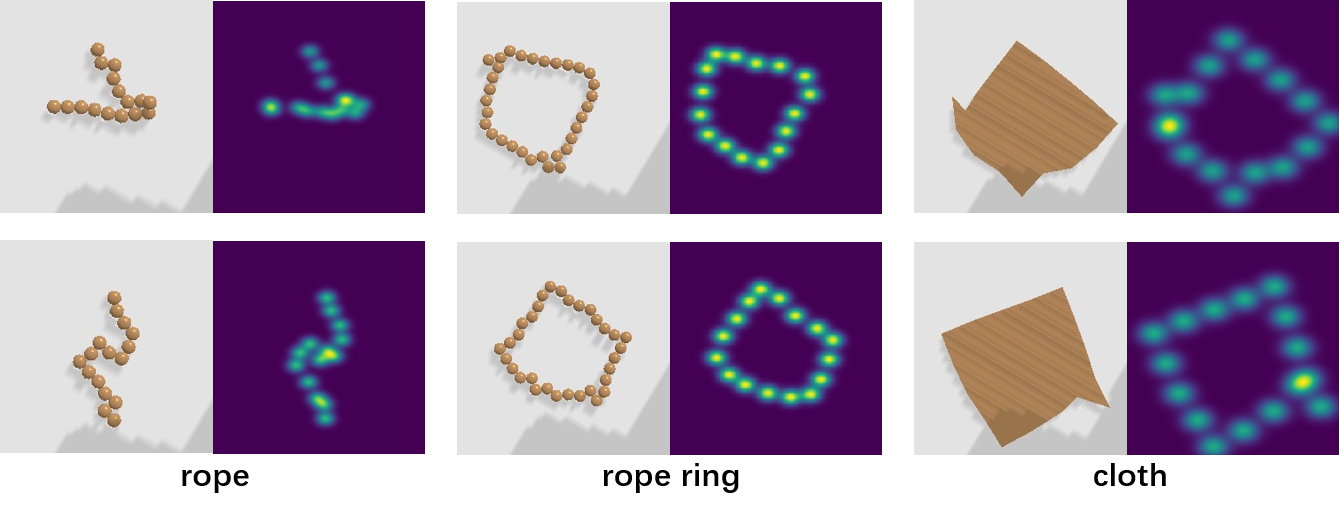}}
\caption{The performance of keypoint detector: our model performs well in detecting keypoints on the rope, rope ring, and cloth. Our model can detect effective points even when there are large-scale occlusions in the image. The left column represents the original image, and the right column represents the gaussian heatmap centered at keypoints detected from the image.}
\label{fig:keypoint}
\end{figure}

The result of keypoint detection is shown in Fig.~\ref{fig:keypoint}. Our keypoint detector perform well in three kinds of deformable object (rope, rope ring, and cloth) involved in our dataset.\par

After getting the current coordinates $C_t$ and the goal coordinates $C_g$ of keypoints, we can get a single complete graphs, whose nodes are $C_t$ and  $C_g$. 

\subsection{Learning for Rearranging Policy}
\begin{figure*}[!t]
\centerline{\includegraphics[width=1\linewidth]{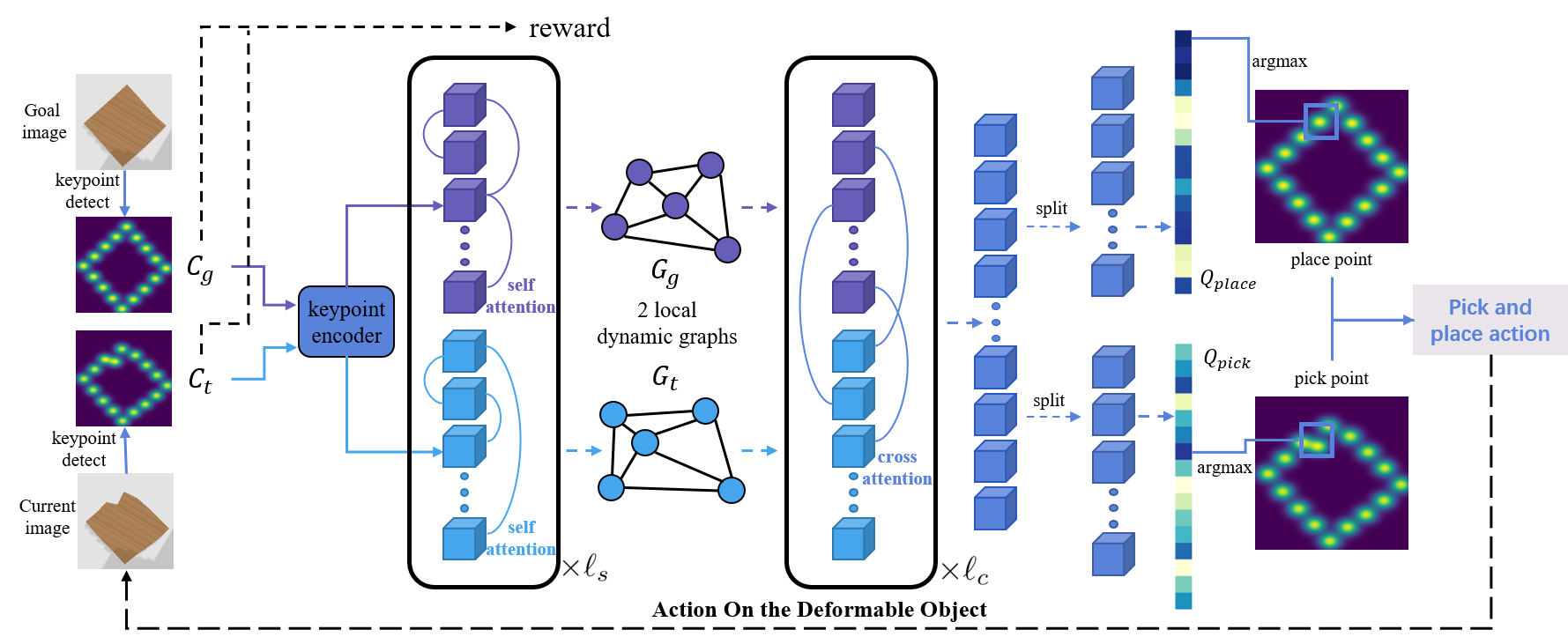}}
\caption{Deep reinforcement learning architecture: our RL agent encodes the representation vectors of keypoints by self-attention layers firstly and gets 2 local dynamic graphs to represent the current and goal configurations. Cross-attention layers are used to map the 2 graphs to the q-value matrix. Coordinates of the picking point are the keypoints in the current image and coordinates of the placing point are the keypoints in the goal image.}
\label{fig:rl}
\end{figure*}

To develop an optimal manipulation policy for the rearranging task, we design a DQN, which aims to generate a manipulation that can narrow the gap between current and goal configurations. The network architecture is shown in Fig.~\ref{fig:rl}.\par 
The classic setting of DQN can be represented as a Markov decision process (MDP) define as a tuple $(S,A,T,r,\gamma,\omega)$, where $S$ is the set of full observable states (representation graphs), $A$ is the set of actions (pick-and-place actions), $T: S\times A \times S$ is the state transition probability function (related to dynamics of the deformable object). The reward function $r(s,a)$ specifies the reward received when the agent takes action $a$ at state $s$, $\gamma$ is the discount factor and $\omega$ denotes all of the parameters of the Q-value network.\par
Given a current state $s_t$, where $t$ is the time instant, the robot could obtain the reward $r_{t+1}=r(s_t,a_{t+1})$, and the accumulated reward is denoted as $R_{t+1} = \sum_{i=t+1}^{\infty}\gamma^{i-t-1}r_i$. For rearranging policy $\pi$, we use $Q^{\pi}(s_t)=\mathbb{E}[R_{t+1}|s_t,\pi]$ to denote the action-value mapping and $Q*$ is the estimated one. We resort to the Bellman equation to calculate the optimal action policy.\par
Considering that $S\in\mathbb{R}^{L\times N}$ ($L=2K$, $N$ is the dimension of node representation vector in the graph) and $A\in\mathbb{R}^{L}$ (we use the position of keypoints in current and goal images as candidate positions for picking and placing), the action-value mapping is a sequence to sequence mapping. So we utilize a GNN (Graph Neural Network) as the function approximator to estimate the action-value mapping $Q(\omega) \rightarrow Q*$.
Parameters $w$ can be iteratively estimated by minimizing the following temporal difference objective:
\begin{equation}
\hat{w} = \mathop{\arg\min}\limits_{\omega} \mathbb{E}[(r_t+\gamma\mathop{\max}\limits_{a_{t+1}}Q(s_t,a_{t+1};\omega)-Q(s_{t-1},a_{t};\omega))^2]
\label{eq_4}
\end{equation}
By formulating the temporal difference error as the following objective:
\begin{equation}
L(w) = (r_t+\gamma\mathop{\max}\limits_{a_{t+1}}Q(s_t,a_{t+1};\omega)-Q(s_{t-1},a_{t};\omega))^2
\label{eq_5}
\end{equation}
We transform the optimization problem into a standard regression estimation problem. We will elaborate the state space, action space, and reward design as follows.

\subsubsection{State Space}
Current and goal coordinates are obtained using keypoint detector developed in Section \ref{sec:graph}, which utilizes the RGB image as the input and produces coordinates of keypoints. We embed the coordinates into a high-dimensional vector with a Multilayer Perceptron (MLP) and get the initial representation $^{(0)}x_p$ for each keypoint $p$. 
\begin{equation}
^{(0)}x_p = {\rm MLP}(c_p)
\label{eq_6}
\end{equation}
We then use a single complete graph whose nodes are the keypoints of current and goal images to construct the state space. The graph is a multiplex graph that has two types of edges: self edges $\mathcal{E}_{self}$ (connect keypoints from the same image) and cross edges $\mathcal{E}_{cross}$ (connect keypoints from the different image).\par
The message passing formulation is used to propagate information along edges. The representation of each point is updated by simultaneously aggregating messages across specified edges. Let $^{(\ell)}x_p$ be the intermediate representation for keypoint $p$ in the single complete graph at layer $\ell$, $\mathcal{E}\rightarrow p$ is the result of aggregation messages from all keypoints $q : (p,q)\in\mathcal{E}$, where $\mathcal{E} \in \mathcal{E}_{self}\cup \mathcal{E}_{cross}$, and  $\left[\cdot\|\cdot\right]$ denotes concatenation. The representation of each keypoint will be updated as follow:
\begin{equation}
^{(\ell+1)}x_p = ^{(\ell)}x_p+\operatorname{MLP}\left(\left[^{(\ell)}x_p \| 
m_{\mathcal{E}\rightarrow p}\right]\right)
\label{eq_7}
\end{equation}
Starting from $\ell=0$, $\mathcal{E}=\mathcal{E}_{self}$ at the first $\ell_s$ layers ($\ell<\ell_s$), where we only aggregate messages across self edges to update the representation vector of each point. In this way, we can establish 2 local dynamic graphs $G_t$ and $G_g$ for current and goal configurations representation at first. After that ($\ell \ge \ell_s$),  $\mathcal{E}=\mathcal{E}_{cross}$ and the messages across cross edges will be aggregated to update the final representation vector for Q-value generation at each point, which can be considered as the matching process between keypoints in $G_t$ and $G_g$.\par
The aggregating message $m_{\mathcal{E}\rightarrow p}$ is computed by an attention mechanism in Transformer~\cite{transformer}, which is widely used in sequence to sequence tasks in Natural Language Processing. The input vectors of each keypoint $p$ for an attention layer is consisted of the value vector $\mathbf{v}_p$, the query vector $\mathbf{q}_p$ that retrieval information from $\mathbf{v}_p$ and the key vector $\mathbf{k}_p$ corresponding to $\mathbf{v}_p$. An attention function can be described as mapping a query vector and a set of key-value vector pairs to an output:
\begin{equation}
m_{\mathcal{E}\rightarrow p} = \sum_{q : (p,q)\in\mathcal{E}}\operatorname{Softmax}_q\left(\mathbf{q}_p^T\mathbf{k}_q\right)\mathbf{v}_q
\label{eq_8}
\end{equation}

\subsubsection{Action Space}
The pick-and-place action is action primitive for rearranging tasks in our solution framework.
To reduce the action space, we have treated keypoints as candidate positions for picking and placing. The keypoints of the current image are candidate positions for picking to eliminate the useless picking on the empty place while the keypoints of the goal image are candidate positions for placing to increase effective placing for rearranging. $a_{t+1}=(pick_{t+1},place_{t+1})$ is the action will be taken at $t+1$, $pick_{t+1}$ is the pick point and $place_{t+1}$ is the place point, $C_t$ and $C_g$ is the current and goal coordinates of keypoints, we have:
\begin{equation}
pick_{t+1} \in C_t \quad place_{t+1} \in C_g
\label{eq_9}
\end{equation}
The design of candidate positions can also formulate our manipulation task as a sequence-to-sequence task, where the input is the initial representation vectors of keypoints of current and goal images and the output is the Q-values of picking and placing on these keypoints.
\subsubsection{Reward Design}
We design the reward function $r(s_{t-1},a_t)$ to encourage effective actions according to the distance between the current and goal coordinates of keypoints on deformable object:
\begin{equation}
r(s_{t-1}, a_t)  = \frac{\left\|C_{t-1}-C_g\right\|_2-\left\|C_{t}-C_g\right\|_2}{\left\|C_{t-1}-C_g\right\|_2}
\label{eq_10}
\end{equation}
where $\left\| \cdot \right\|_2$ denotes Euclidean distance. We normalize the distance change after the action $a_t$ is executed by dividing it by the distance at the $t-1$ instant.

\begin{figure}[!t]
  \centering
    \subfigure[rope]{\includegraphics[width=0.32\linewidth]{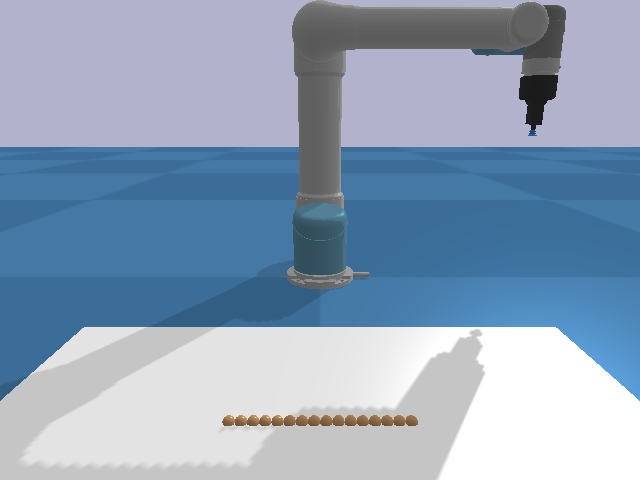}}
    \subfigure[rope ring]{\includegraphics[width=0.32\linewidth]{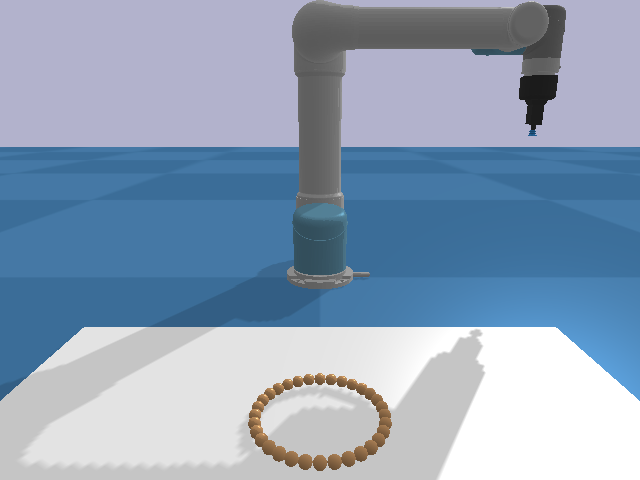}}
    \subfigure[cloth]{\includegraphics[width=0.32\linewidth]{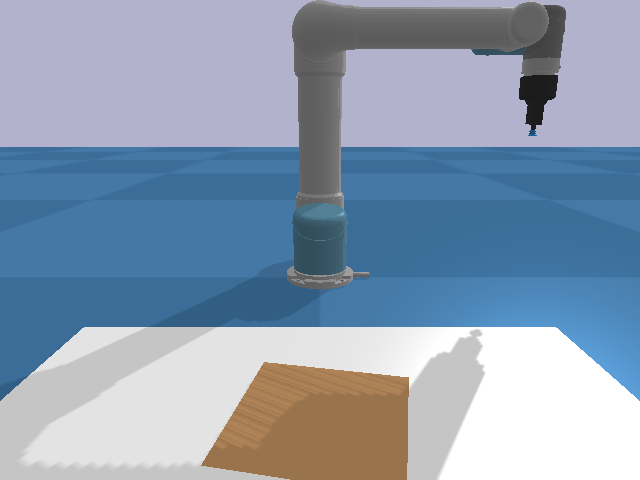}} \\
  \caption{Simulation environment: we resort to the pybullet robotic engine to construct simulation environments of three kinds of deformable objects involved in our dataset.}
    \label{fig:data_scene}
    \vspace{0.2in}
\end{figure}

\section{Dataset}
\label{sec:dataset}
\begin{figure}[ht]
\centerline{\includegraphics[width=1\columnwidth]{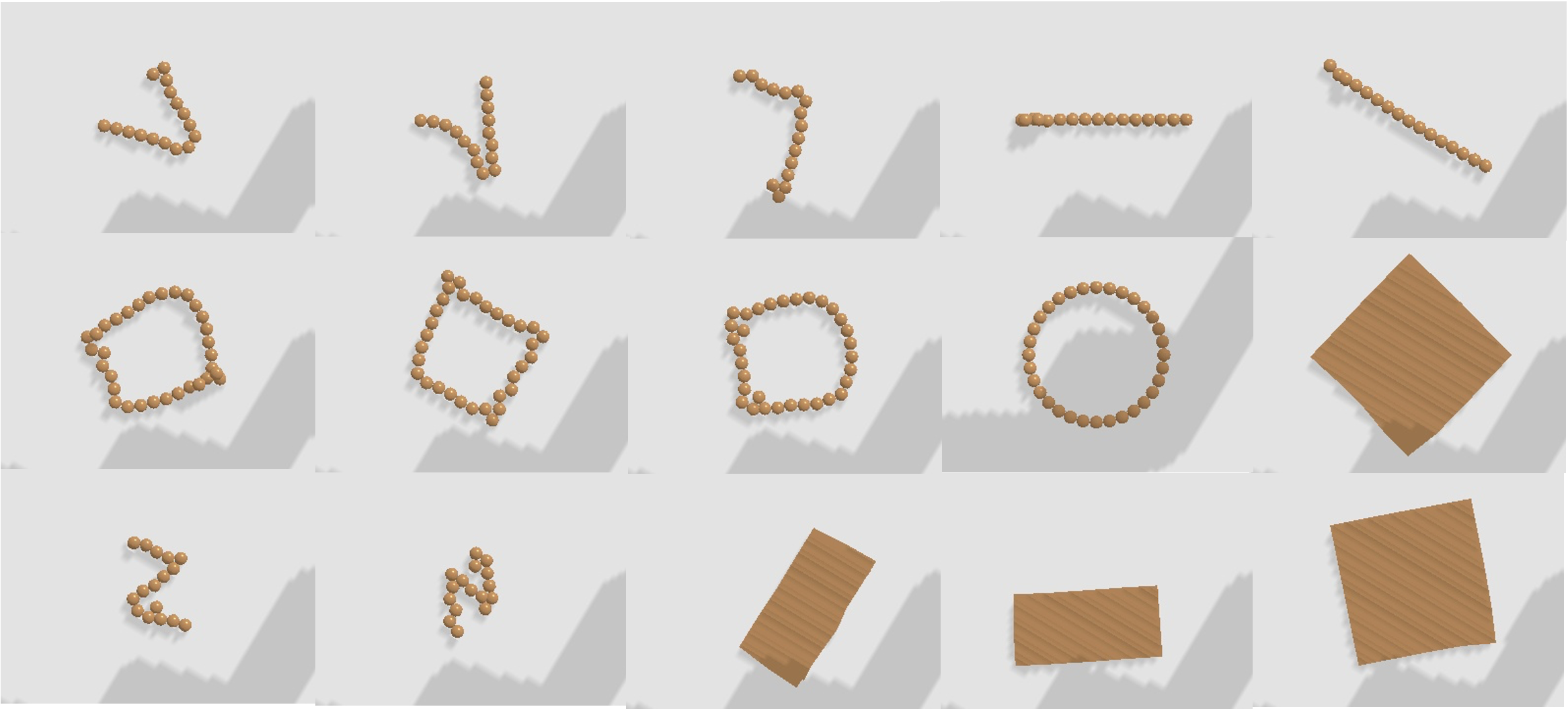}}
\caption{The goal configurations in our dataset are random. The orientation, the length ratio of sides, the included angle, and the position are not specific but given in the goal image.}
\label{fig:goal}
\end{figure}
Real robotic experiments are usually not scalable for the restricted experimental environment and the number of the robotic dataset collected in the real environment is limited by safety and cost aspects, which is not enough for the training of robotic algorithms. So we modify the dataset in~\cite{transporter_df} and establish our dataset of deformable object rearranging tasks in the pybullet~\cite{pybullet} simulation environment. Pybullet robotic engine can provide realistic visual renderings and support for 1-D and 2-D deformable object simulation. We put a single camera in a top-down view of the planar surface and construct a dataset for our rearranging tasks.\par

As Fig.~\ref{fig:data_scene} shows, our dataset involves three different types of deformable objects (rope, rope ring, and cloth) each with multiple related rearranging tasks, which makes the types of tasks in our dataset sufficient to verify the generalization and effectiveness of the framework. The rearranging tasks involved in our dataset are shown in the TABLE~\ref{table:dataset}.\par
\begin{table}
\caption{Rearranging tasks involved in our dataset}
\label{table:dataset}
\setlength{\tabcolsep}{4pt}
\centering
\begin{tabular}{|p{40pt}|p{40pt}|p{140pt}|}
\hline
Dimension& Categories& Tasks\\
\hline
1D& rope& rope straightening \\
1D& rope& rearrange the rope into a V-shape \\
1D& rope& rearrange the rope into a N-shape \\
\hline
1D& rope ring& rearrange the rope ring into a circle  \\
1D& rope ring& rearrange the rope ring into a square  \\
1D& rope ring& move the rope ring to a random position \\
\hline
2D& cloth& cloth flattening \\ 
2D& cloth& cloth folding  \\
\hline
\end{tabular}
\label{tab1}
\end{table}
It is worth noting that the goal configurations in our dataset are random rather than specific (Fig.~\ref{fig:goal}). The shape (the length ratio of two sides and the included angle of the V-shape, the included angle of the N shape), the orientation, and the position of the goal configurations are all random. Robots need to learn more effective rearranging skills.  

\begin{figure*}[!t]
\centerline{\includegraphics[width=1\linewidth]{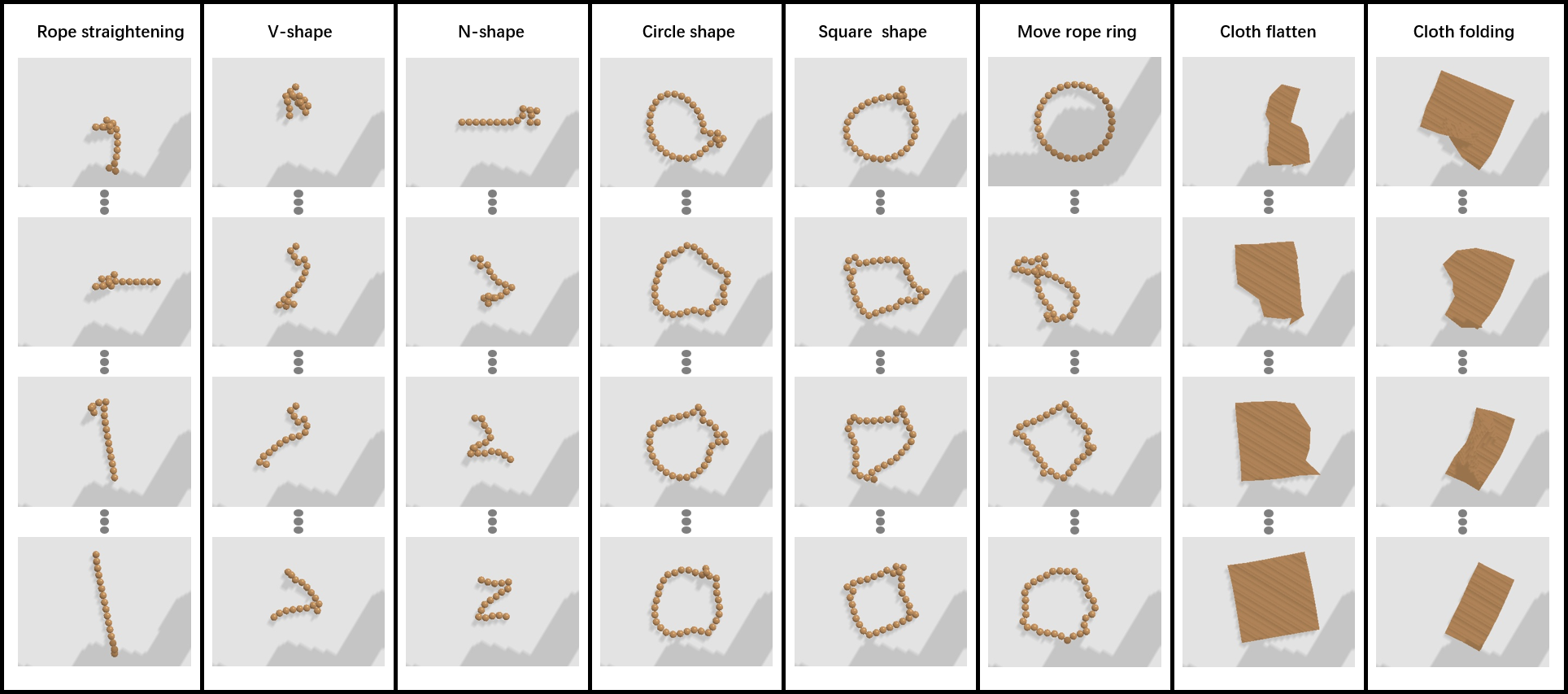}}
\caption{We evaluate our framework on 8 types of deformable object rearranging tasks involved in our dataset. Each example shows four frames in the sequence of a rearranging task. Experimental results show that our framework is general and effective.}
\label{fig:vali}
\end{figure*}
\section{Experiment Result}
\label{sec:experiment}

This section presents experiments to show the performance of our method.

\begin{table}[!t]
\centering
\caption{Evaluation results of tasks involved in our dataset}
\setlength{\tabcolsep}{4pt}
\begin{tabular}{|p{150pt}|p{70pt}|}
\hline
 Tasks &Successful rate\\
\hline
rope straightening& 100\% \\
rearrange the rope into a V-shape& 98\% \\
rearrange the rope into a N-shape& 75\% \\
\hline
rearrange the rope ring into a circle& 72\%  \\
rearrange the rope ring into a square& 100\%  \\
move the rope ring to a random position& 70\% \\
\hline
cloth flattening& 90\%  \\ 
cloth folding& 86\%   \\
\hline
\end{tabular}
\label{tab:experi}
\end{table}

\subsection{Evaluation Experiment in Simulation}
We first conduct experiments to evaluate the performance of our framework on multiple rearranging tasks. The robot is given random goal configurations by only visual input, and the robot needs to rearrange the deformable object to the goal configurations without any sub-goal input. We define that the robot completes a rearranging task within 30 picking and placing actions as a success, and the rest is a failure. The situation of completing a rearranging task is that the average pixel distance between the corresponding keypoints in the current and goal image is less than 10. We have chosen this threshold through lots of experiments. The two configurations that satisfy this situation are similar enough. We evaluate the model on the new 100 testing goal-conditioned rearranging tasks for each type of task. The result of the evaluation experiment is shown in TABLE~\ref{tab:experi}. The tasks involved in the circle shape (the geometry is very strict) are hard to complete for complex dynamics, the success rates have reached 75\% (rearranging into a circle) and 70\% (move the rope ring) even on these tasks. The success rate is also acceptable on the 2-D deformable object rearranging task. Compared with the previous framework for multi-task deformable object rearranging skill learning (the tasks involved are different)~\cite{transporter_df}\cite{lim2021multi}, our success rate is almost equivalent, which can verify the performance of our proposed model.\par

\begin{figure}[!t]
\centerline{\includegraphics[width=1\columnwidth]{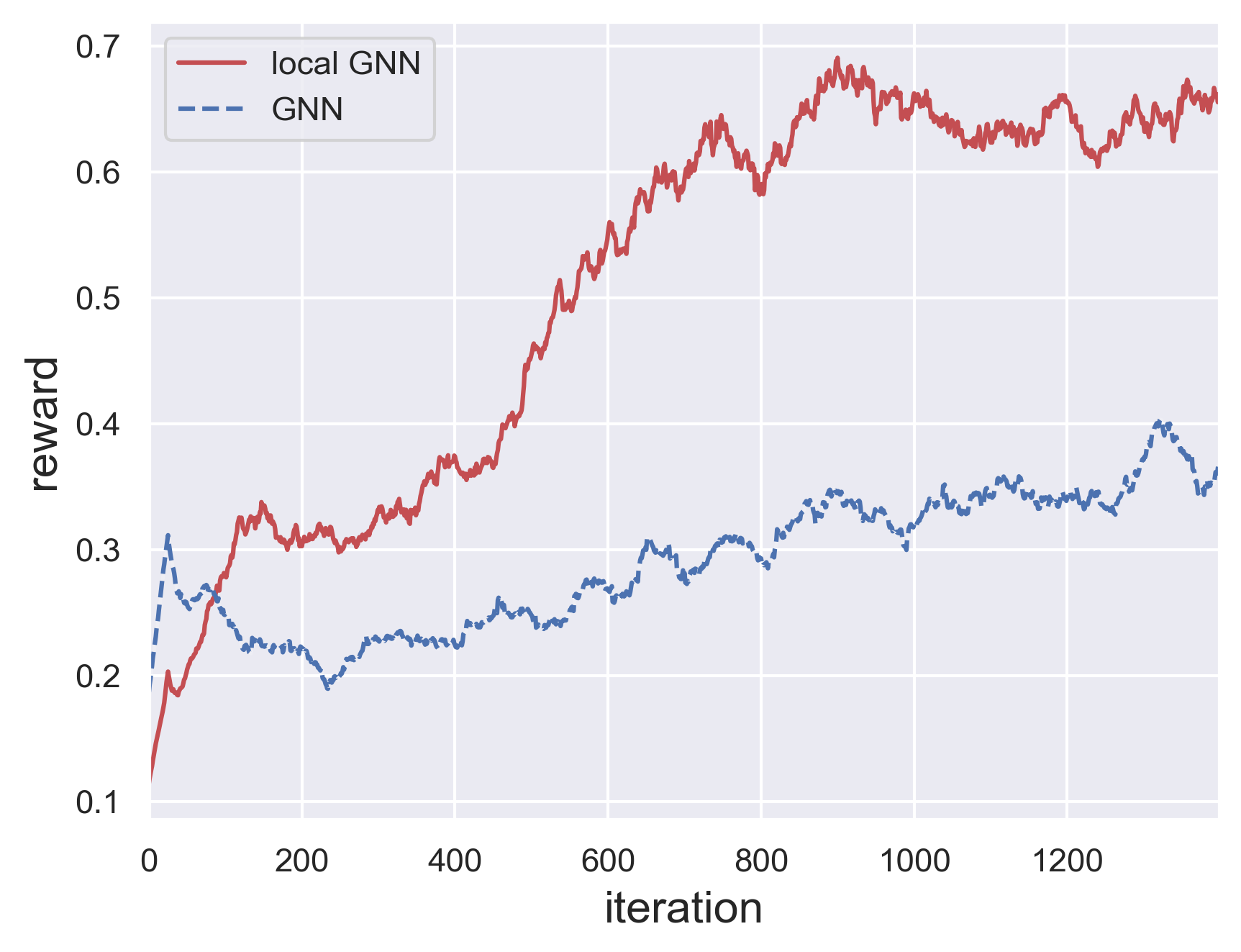}}
\caption{The reward obtained of GNN and local GNN.}
\label{fig:reward}
\end{figure}

However, the training of our model is much faster and the size of our model is much smaller. Because we don't rely on the CNN-oriented image feature adopted in previous work, our graph feature is better at handling the sparse information of the deformable object. The calculation of attention on short sequences (32 keypoints in our model setting) is more efficient than multiple convolution calculation on the whole image, where the most area without the deformable object is redundant. Fig.~\ref{fig:vali} shows some examples of goal-conditioned deformable object rearranging tasks involved in our dataset. All of these tasks are completed in 30 manipulations.\par

\subsection{Ablation Study of Local GNN Architecture}

\begin{figure*}[th]
\centerline{\includegraphics[width=1\linewidth]{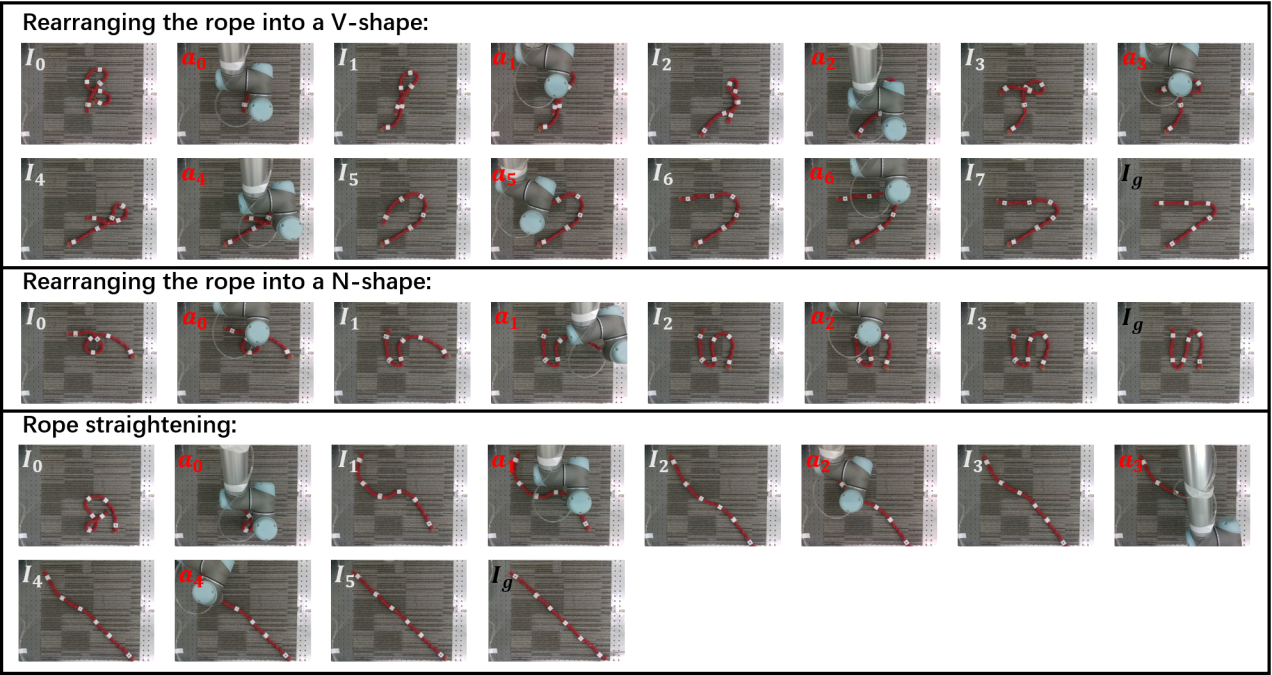}}
\caption{The robot completes the rope straightening, rearranging the rope into a V-shape and an N-shape in reality.}
\label{fig:real_result}
\end{figure*}
Instead of directly performing global attention calculation on the sequence of keypoints to generate the distribution of the Q value, we have adopted a local GNN. An experiment has also been conducted to validate the rationality of this design. We compared the reward obtained of our local GNN and the origin GNN in the rearranging task training. The number of layers for both models is set to be the same. As Fig.~\ref{fig:reward} shown, the reward obtained by local GNN is larger than the reward obtained by origin GNN. The improvements convincingly demonstrated our model design's positive contribution to agent performance. At the same time, the computational cost of local GNN is half that of origin GNN because the attention calculation is only performed on points from the same image (self-attention) or points from the different image (cross-attention) at each layer, which can further improve the efficiency of our model.

\subsection{Physical Experimental Validation}
\begin{figure}[th]
\centerline{\includegraphics[width=0.9\columnwidth]{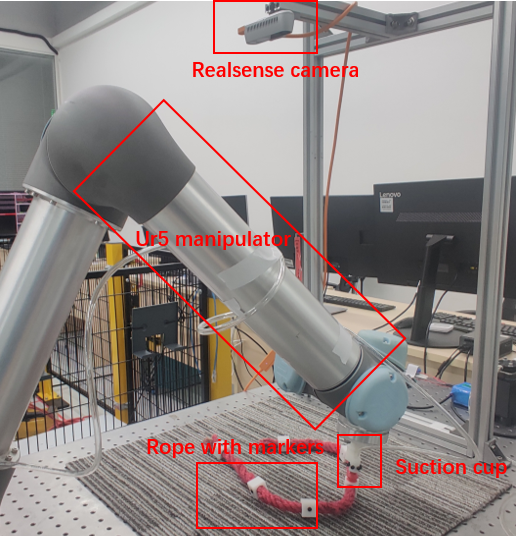}}
\caption{Our physical experimental setup consists of a UR5 manipulator, a Realsense camera, a suction cup, and a rope with markers.}
\label{fig:real_scene}
\end{figure}
In our framework, the processing of visual features (keypoint detector) and the planning of manipulation (local GNN) are naturally separated, which enables the transfer of our framework from simulation to reality. Our model can learn multiple goal-conditioned deformable object rearranging skills from a large quantity of data in simulation and these skills can be used in reality with only keypoint detector fine-tune. We test our proposed method in a physical environment with the setup shown in Fig.~\ref{fig:real_scene}. The rope is placed on the platform, and a UR5 robotic manipulator with a suction cup is placed in front of the platform for picking and placing. Images are captured with a Realsence camera, which is fixed on the top of the platform. It's should be noticed that we use a suction cup as the end-effector and we add 6 markers to attach to the rope for sucking. These markers have litter influence on the dynamic of the rope but can make our experiment easier to conduct. It's easy to replace the sucker with the gripper by considering orientation in the picking and placing action. The orientation can be the tangent direction at the picking and placing point. \par
Affected by the Covid-19 epidemic, we only tested the goal-conditioned rope rearranging task. We collect 500 real images of the rope and fine-tune the keypoint detector to reduce errors caused by image style differences. The number of keypoints is set as 6, which is different from the simulation. Our model relies on GNN, which makes the number of keypoints can be not fixed. We can choose a different number of keypoints for deformable objects with different physical properties in reality, which further ensures the transferability of the model. The results are shown in Fig.~\ref{fig:real_result}. $I_t$ denotes the image at $t$ instant, $I_g$ denotes the goal image of the task and $a_t$ denotes the picking and placing action taken at $t$ instant. The experimental results demonstrate that the skills our framework learned in the simulation are also effective in reality.

\section{Conclusion}
\label{sec:conclusion}

In this paper,  we propose a novel framework to solve the task of goal-conditioned deformable object rearranging. The configurations of the deformable object are represented as a dynamic graph in our model design. The graph feature is passed through a local GNN to obtain Q-value distribution in reinforcement learning. Extensive experiments have been conducted demonstrating that the rearranging policies learned by our framework not only perform well in simulated scenarios but can also be transferred to real scenarios with only keypoint detector fine-tune.


\end{document}